%% file: main.tex
\documentclass[conference]{IEEEtran}

\usepackage[english]{babel}
\usepackage[utf8]{inputenc}
\usepackage[T1]{fontenc}

\usepackage[dvipsnames]{xcolor}

\usepackage{amsmath}
\usepackage{amssymb}
\usepackage{amsfonts}

\makeatletter
\newcommand*{\transpose}{{\mathpalette\@transpose{}}}
\newcommand*{\@transpose}[2]{\raisebox{\depth}{$\m@th#1\intercal$}}
\makeatother

\usepackage[binary-units=true]{siunitx}
\usepackage{dcolumn}
\usepackage{verbatim}

\usepackage[colorlinks]{hyperref}

\usepackage{caption, subcaption}

\usepackage{prettyref}
\newrefformat{eq}{Equation (\ref{#1})}
\newrefformat{lem}{Lemma \ref{#1}}
\newrefformat{thm}{Theorem \ref{#1}}
\newrefformat{cha}{Chapter \ref{#1}}
\newrefformat{sec}{Section \ref{#1}}
\newrefformat{tab}{Table \ref{#1}}
\newrefformat{fig}{Figure \ref{#1}}

\usepackage{tikz}
\usepackage{standalone}
\usepackage{pgfplots}
\newlength\fwidth
\usetikzlibrary{plotmarks}
\usetikzlibrary{arrows.meta}
\usetikzlibrary{spy}
\usepgfplotslibrary{patchplots}
\usepackage{grffile}
\pgfplotsset{compat=newest}

\begin{document}
\title{The Autonomous Racing Software Stack of the KIT19d}

\author{%
\IEEEauthorblockN{Sherif Nekkah\IEEEauthorrefmark{1}, Josua Janus\IEEEauthorrefmark{1}, Mario Boxheimer\IEEEauthorrefmark{1}, Lars Ohnemus\IEEEauthorrefmark{1}, \\
Stefan Hirsch, Benjamin Schmidt, Yuchen Liu, David Borbély, Florian Keck, \\ 
Katharina Bachmann, Lukasz Bleszynski\vspace{2mm}}
\IEEEauthorblockA{\textit{KA-RaceIng e.V.}\\
\textit{Karlsruhe Institute of Technology, Karlsruhe, Germany}\\
\textit{Mail: \{prename\}.\{surname\}@ka-raceing.de}\\
\IEEEauthorrefmark{1}Authors with equal contribution}}

\maketitle
\input{Sections/01_Abstract}
\input{Sections/02_Introduction}
\input{Sections/03_Related_Work}
\input{Sections/04_Methods}
\input{Sections/05_System_Architecture}
\input{Sections/06_Perception}
\input{Sections/07_Localization_and_Mapping}
\input{Sections/08_Motion_Planning_and_Control}
\input{Sections/09_Results}

\input{Sections/10_Conclusion}

\bibliographystyle{abbrv}
\bibliography{paper_bibtex}

\end{document}

%% file: Sections/01_Abstract.tex
\begin{abstract}
Formula Student Driverless challenges engineering students to develop autonomous single-seater race cars in a quest to bring about more graduates who are well-prepared to solve the real world problems associated with autonomous driving. In this paper, we present the software stack of KA-RaceIng's entry to the 2019 competitions. We cover the essential modules of the system, including perception, localization, mapping, motion planning, and control. Furthermore, development methods are outlined and an overview of the system architecture is given. We conclude by presenting selected runtime measurements, data logs, and competition results to provide an insight into the performance of the final prototype.
\end{abstract}

%% file: Sections/02_Introduction.tex
\section{Introduction}

The third Formula Student Driverless (FSD) competition was held at the Hockenheimring in Germany from the 5th to the 11th of August 2019. The competition was introduced in 2017 and extended the previously existing combustion and electric classes. Since then, KA-RaceIng\footnote{\url{https://www.ka-raceing.de}}, the Formula Student team of the Karlsruhe Institute of Technology\footnote{\url{http://www.kit.edu}} (KIT) is competing in all three classes. Meanwhile, the driverless series has become a research platform for cutting edge technology in the area of autonomous driving. 

\begin{figure}
  \centering
  \includegraphics[width=.99\linewidth]{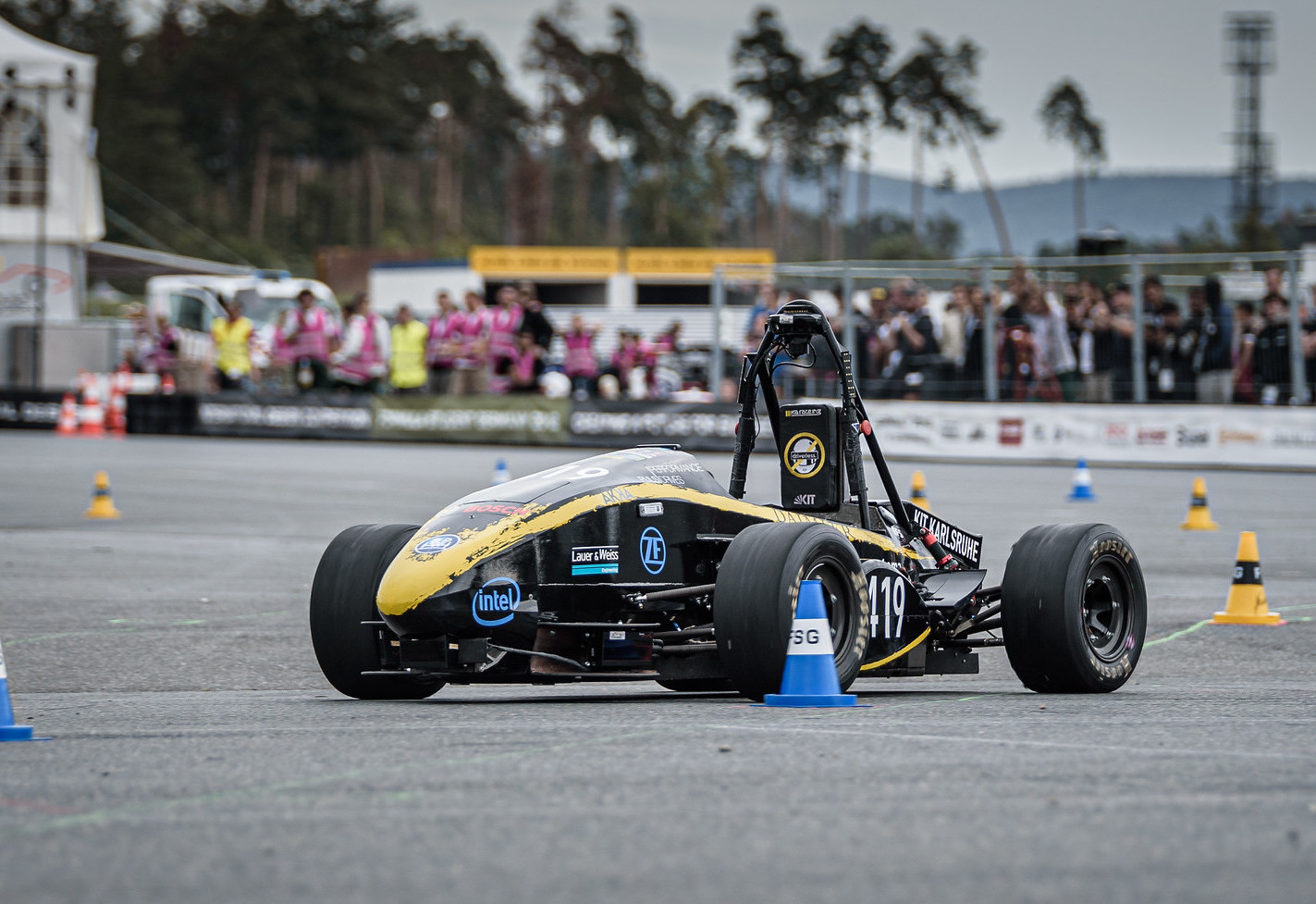}
  \caption{\small{The KIT19d. Autocross, Trackdrive and Skidpad winner in Formula Student Germany 2019. First place overall in Formula Student Spain 2019. Photo by Zenker, ©FSG.}}
  \label{fig:KIT19_FSG}
\end{figure}

The competition consists of four static and four dynamic disciplines \cite{fsg2019rules}. Static disciplines challenge the student teams beyond the development of an autonomous race car and evaluate their knowledge in terms of hypothetical business and cost plans as well as their engineering know-how. The dynamic disciplines test the vehicle's performance and reliability under high longitudinal and lateral accelerations, as well as the system's ability to race on unknown tracks. As shown in \prettyref{fig:KIT19_FSG}, the vehicles race without a human fallback driver. The boundaries of the race track are defined by yellow traffic cones to the right and blue ones to the left, which must be identified autonomously by the system.

This paper introduces the autonomous system (AS) software design of the KIT19d\footnote{\url{https://www.ka-raceing.de/19d}}. Arguably the most challenging discipline of the FSD competition is the Autocross, in which the vehicle has to complete a complex and unknown course as fast as possible. We believe that a vehicle that is competitive in the Autocross event will also be competitive in the other dynamic disciplines. Therefore, much attention was directed at the performance of the system in the Autocross event. See \url{https://youtu.be/sxqkt\_ydOkY?t=3155} for the run at the Hockenheimring in Germany. Furthermore, \url{https://youtu.be/h22J8YzNdjo} provides a visualization of the mapping and planning process during this run. 

At the beginning of the project, the following main goals for the AS software have been set:

\quad\textit{Modularity} enables the development of a well-structured software architecture that is a sustainable base for future competition seasons.

\quad\textit{Reliability}
is the major goal behind each concept and design decision as it is the key to success in the Formula Student Driverless competition.

\quad\textit{Efficient Design} 
enables a small team to develop a functional AS despite limited resources and leads to a lightweight system that is easily surveyed and tested.

\quad\textit{Performance}
improvement is the driving force behind most new developments. Provided these developments do not deteriorate reliability, they contribute to high scores in static and dynamic disciplines. \\

This paper is structured as follows: \prettyref{sec:methods} outlines the development methods we found to work well in the context of the limited resources of Formula Student teams. Sections \ref{sec:sys_architecture} - \ref{sec:motion_planning} present the technical features of the KIT19d's AS where \prettyref{sec:sys_architecture} covers the hardware and software architecture. \prettyref{sec:perception} presents the perception module, \prettyref{sec:slam} outlines localization and mapping, and \prettyref{sec:motion_planning} describes methods used for motion planning and control. Performance evaluations of the resulting system can be found in \prettyref{sec:results} and conclusions and an outlook are provided in \prettyref{sec:conclusion}.

%% file: Sections/03_Related_Work.tex
\section{Related Work}
Several Formula Student teams have published overview papers, describing the software stack developed for their autonomous vehicles, such as the teams from Zurich \cite{valls2018design, kabzan2019amz}, Vienna \cite{zeilinger2017design}, and Beijing \cite{jun2018autonomous}. Autonomous racing cars have also been developed for other competitions, including the DARPA Grand Challange \cite{thrun2006stanley}, Roborace \cite{heilmeier2019minimum} and the Carolo Cup \cite{nolte2018carolo}. For the sake of brevity we omit a deeper literature review here, but we will point to relevant publications for the methods we use in the following sections.

%% file: Sections/04_Methods.tex
\section{Methods}\label{sec:methods}

In Formula Student, race cars are developed in less than a year. Thus, strategies to evaluate concepts rapidly and efficient methods for testing and validating the results are the keys to fast improvements. However, most models employed in larger companies are not compatible with a Formula Student team structure. A limited amount of team members, time, and the lack of experience require workflows that can be adapted in close to no time and do not produce a lot of additional workload. To ensure flexibility in the task assignment, our software development process contained elements from SCRUM \cite{SCRUM}. Weekly reviews allowed constant tracking of progress and, if unavoidable, the relocation of resources. 

When going into a new competition season, the first task is to identify the modules that need to be improved and to allocate the resources to do so. Reviewing previous development cycles, vulnerabilities can be determined and addressed. Testing processes must be tailored for all components whilst balancing their complexity and the level of system integration. More specifically some modules require complete simulations while others can be tested with recorded data. For the development of the motion planning and control module, for instance, we considered a simulation to be necessary. In this case, the simulation is required to generate a feedback loop around a model that is an accurate representation of the real vehicle behavior. Existing vehicle dynamic models were used and expanded with the required interfaces of the AS. Based on rviz\footnote{3D visualization tool from the open-source framework ROS \cite{Quigley2009ROSAO}}, a 3D visualization of the resulting vehicle trajectory facilitates the interpretation of simulation results and allows for fast parameter evaluation and tuning. In contrast, for the perception, the localization, and the mapping modules, only replays of recorded data from real sensors were used for debugging purposes instead of a full model-based simulation. For these purposes, it is very demanding to create a simulation that is as accurate as real data. A drawback of this approach is that the system integration only happens on the vehicle. In general, we want to emphasize that these test routines were an important prerequisite for the success of our developments. Test environments for all subsystems should be developed early. Besides being user-friendly, these should have well-defined interfaces to ensure repeatability and to avoid brute-forcing solutions. 

Past experiences have shown, that coding errors and bad coding practices lead to delays in the deployment process. Consequently, a proper git workflow including code style rules, reviews, and automated tests is recommended.

%% file: Sections/05_System_Architecture.tex
\section{System Architecture}\label{sec:sys_architecture}

\begin{figure*}
  \centering
  \includegraphics[width=.99\textwidth]{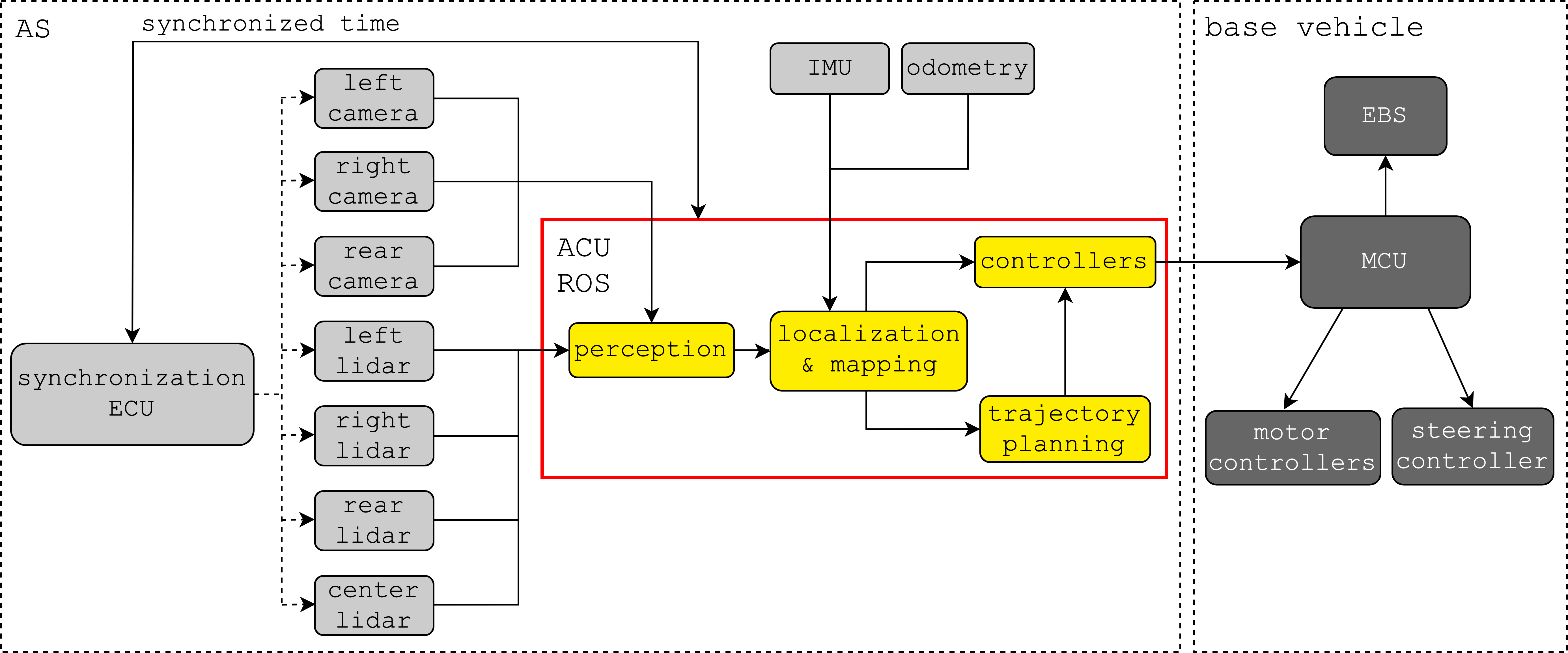}
  \caption{\small{Overview of the AS architecture of the KIT19d.}}
  \label{fig:ASoverview}
\end{figure*}

The hardware platform is provided by the KIT15e\footnote{\url{https://www.ka-raceing.de/15e}}, an all-wheel-drive electric vehicle. It was retrofitted with the necessary components for autonomous racing, such as an emergency brake system (EBS), four lidars, three cameras, and a steering actuator. The driver, who would normally provide the control inputs was replaced by an autonomous system control unit (ACU), that perceives the environment, plans an appropriate trajectory, and controls the longitudinal and lateral motion of the vehicle. 

\subsection{Overall Pipeline}

\begin{figure}[ht]
  \centering
  \includegraphics[width=.99\linewidth]{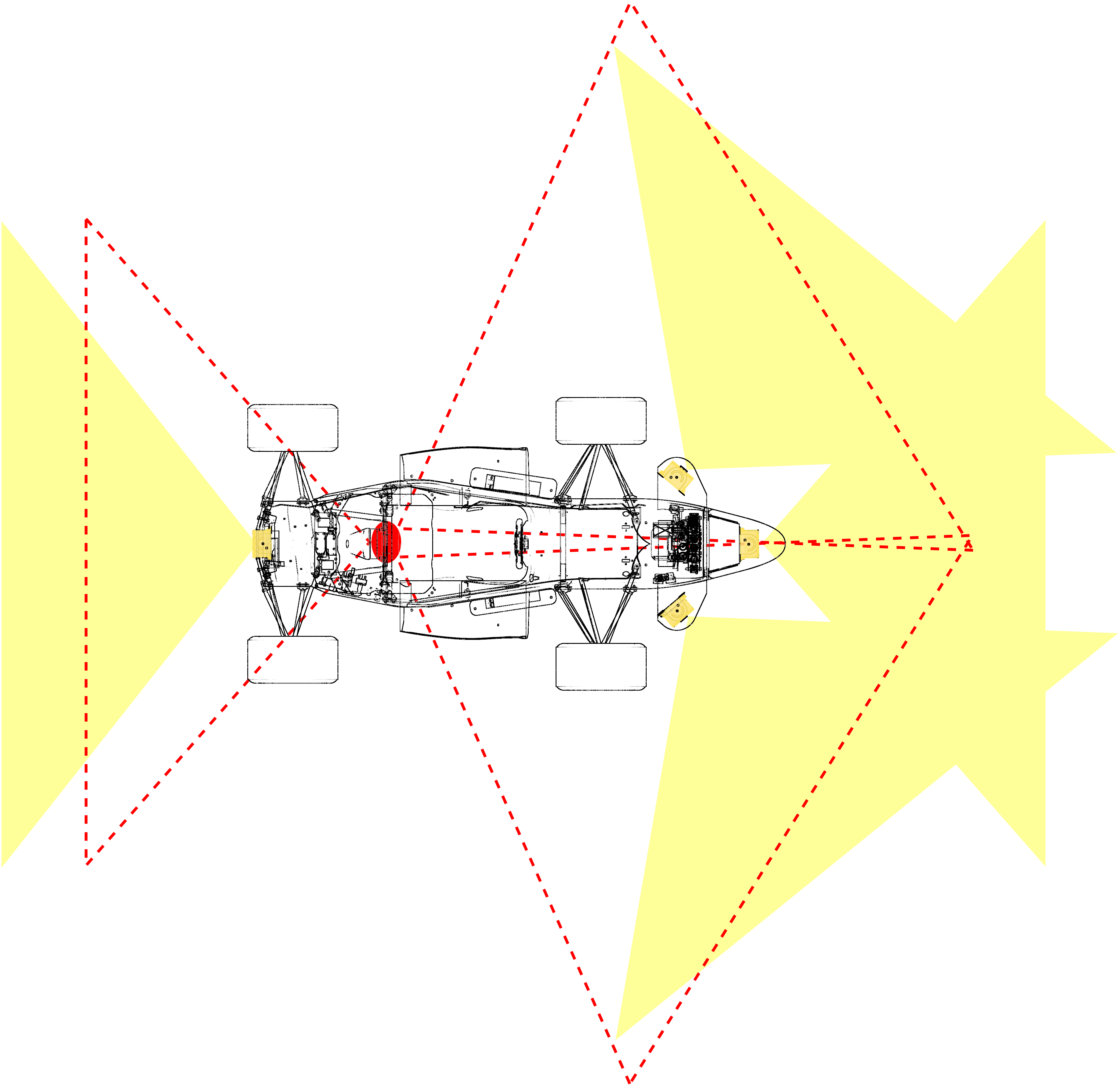}
  \caption{\small{Field of view of the cameras (red) and the lidars (yellow) of the KIT19d.}}
  \label{fig:fow}
\end{figure}

An overview of the complete autonomous system is given in \prettyref{fig:ASoverview}. To acquire extensive information about the track layout early-on, the car is equipped with both forward and rearward facing cameras and lidars, as shown in \prettyref{fig:fow}. This enables the detection of objects at ranges up to \SI{42}{\meter} around the vehicle to provide adequate information for the mapping and path planning. To complement the information generated by the perception system, an inertial measurement unit (IMU) and wheel speed sensors were added to the car. The output of these sensors is forwarded to the localization and mapping algorithm, where the data is combined with detected objects from the perception system and is used to estimate the vehicle pose and to create a map of the environment. During a fully autonomous race on an unknown track, as in the Autocross event, the desired vehicle path is continuously planned on a growing map.

The planned path and the estimated vehicle state are provided to the longitudinal and lateral controllers, which calculate the desired motor torques and the steering angle. The main control unit (MCU) provides the interface between the AS and the electronic infrastructure of the base vehicle. It controls all low-level features of the car, such as brake lights and fans, implements all safety checks necessary for rules compliant operation, and forwards the torque and steering request to the inverter and the steering controller.

\subsection{Autonomous Control Unit (ACU)}

We decided to use the robot operating system (ROS) \cite{Quigley2009ROSAO} on our ACU, to manage the increasing complexity of the architecture, and to follow the goal of a modular structure. This allows for rapid advancements in the overall system by exchanging single blocks with enhanced versions as all interfaces are defined in the beginning of the development process. ROS also facilitates software deployment by providing numerous tools for debugging, system analysis, visualization, recording as well as replaying sets of data. The software stack was designed to run on Ubuntu 18.04, within the ROS melodic framework. The ACU is comprised of commercially available consumer-grade computer hardware, specifically, an Intel i7-9700k CPU, an ASRock x370 Mini-ITX mainboard with \SI{32} {\giga\byte} DDR4 RAM, a Samsung 970 EVO solid-state drive (SSD) and a CAN-Interface to communicate with the MCU. Note that we do not require a dedicated GPU or TPU. Three cameras are attached to the ACU via USB 3.0. The four lidars are connected via Ethernet. 

\subsection{Synchronization}

To exploit the full potential of our lidar concept, the point clouds of the individual lidars have to be merged before they are further interpreted. Therefore, the scans must be started simultaneously. For this purpose, we have developed a control unit (sync-ECU) that provides a common time base to the lidars by generating a synchronization pulse. It also allows us to trigger them asynchronously, which effectively doubles the sample rate in the areas where the field of view of two lidars overlaps. The sync-ECU and the ACU synchronize their clocks using an adapted version of the precision time protocol (PTP) \cite{ptp} over CAN. The four lidars are connected to the ACU via Ethernet, which introduces a significant but non-deterministic latency. When a new scan is received by the ACU, it must therefore first be assigned to the other scans from the same time step. Once the system is initialized, the correct assignments can be determined comparing the scan counter of the individual lidars. However, the counters can differ at power-up and this difference has to be determined upon initialization. This is done by searching for a series of consecutive scans where the latency of each scan is below the empirically determined average latency of the system. When a series is found, the counter offsets are calculated. Note that this algorithm assumes that the average latency is lower than the scan rate of the lidar. The latency can be determined using the timestamps of the start of a scan published by the sync-ECU and the time recorded by the ACU when receiving a scan.


%% file: Sections/06_Perception.tex
\section{Perception}\label{sec:perception}
Intuitively, race cars have to act in a highly complex environment. Navigating the race track requires knowledge about the car's environment, that has to be perceived "on the fly". Building such a perception system includes the detection of relevant features to perceive the track's borders. For a Formula Student race, an obvious choice are the traffic cones used to confine the race track. Since all of these features can be considered static, their location relative to the car's position can be used to subsequently localize the car within the track. 

Our system uses a variety of sensors for perception. The most important ones are multiple monocular cameras and lidars. Lidar sensors create precise but sparse range measurements while camera sensors capture dense image information similar to the human eye. By using a combination of both sensors, we can accurately detect landmarks with high confidence. This section discusses the use of both the lidar and camera pipeline to locate traffic cones, their fusion, and interfaces between the car's perception and other subsystems.

\subsection{Overall Perception Pipeline}
\begin{figure}
    \centering
    \includegraphics[width=\linewidth]{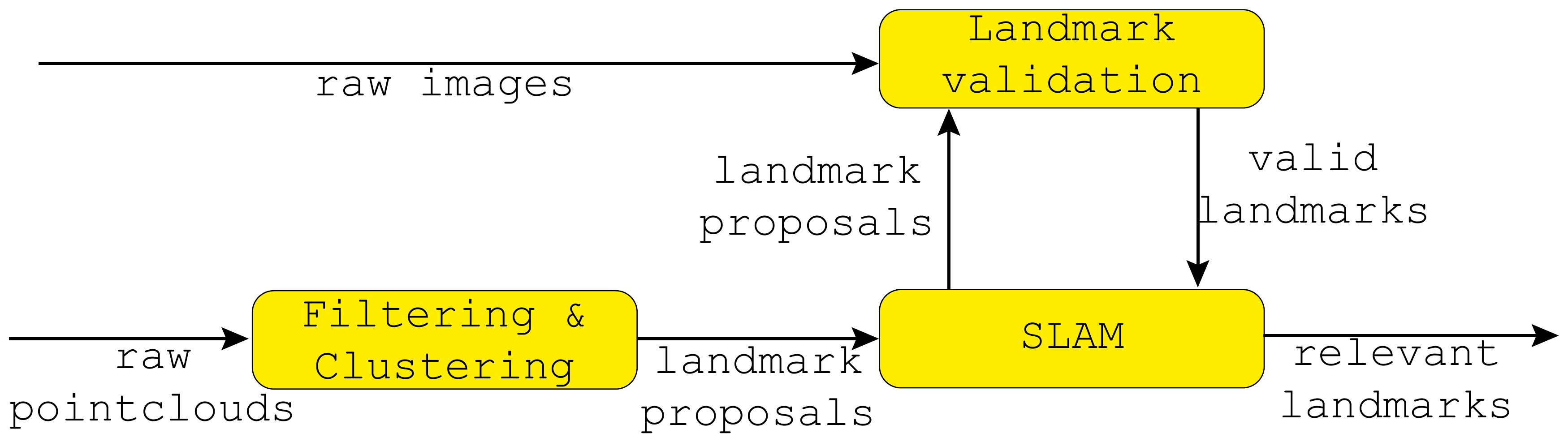}
    \caption{\small{Overall perception pipeline: Lidar range measurements in the form of 3D point clouds and camera images are fed into the perception system. The lidar points are used to generate landmark proposals that are tracked in a global map (see \prettyref{sec:slam}). All proposals are validated against the camera images.}}
    \label{fig:perception}
\end{figure}
In principle, landmarks can be detected with either one of both sensor systems. This creates a robust system because each subsystem works independently and extracts the same information targets. Such multi-sensor setups rely heavily on the accuracy of all subsystems to achieve good results since all measurements are passed on without external validation. As an improvement, all sensor outputs are cross-validated against other systems. Even fusion of raw sensor data is possible in some cases. Localization through the detection of landmarks requires a precise position estimate of these landmarks. In contrast to lidar, this is hard to achieve with monocular cameras.  Here, a projection from the two-dimensional image plane into 3D space has to be done. A homography between the image plane and the ground plane of the track can be calculated, but the accuracy greatly decreases with the distance between sensor and object, due to nonlinear distortions and discretization of positional information.  

We bypassed this issue by restricting the information retrieval of the cameras to their respective image planes. We use accurate lidar detections to generate proposals for landmarks. These proposals are passed to the mapping algorithm (\prettyref{sec:slam}) to store them on a global map. This enables a reliable tracking of multiple landmarks. Landmark proposals are then validated inside the respective camera image. Therefore, the camera sensors work as a validation-device. Furthermore, the camera information is used to infer additional information such as the color of the landmarks. The overall pipeline is depicted in \prettyref{fig:perception}.
Using such an architecture has multiple advantages. We combine precise position measurements given by the lidar sensors with abstract texture and color information of the cameras. Not relying on positional information of the cameras increases the overall system efficiency since no complex object detection techniques need to be applied. Therefore all of the computations can be performed on a standard CPU, which removes the need for an expensive, power consuming GPU. Validating proposals in pixel-space also removes the necessity of a projection to global coordinates. 

\subsection{Landmark Proposals}
    As outlined above, we use only the lidar sensors to generate landmark proposals.\\
The lidars capture a set of $N_{\text{raw}}$ measured points 
\begin{equation}
    P=\left\lbrace \mathbf{p}_i=\left(x_i,y_i,z_i\right)^\transpose \: | \: i=1,2,\ldots,N_{\text{raw}}\right\rbrace,
\end{equation}
called a point cloud $P$. A series of transformations is applied to transform all lidar measurements into one frame of reference.  We can state that only a subset of $P$ actually belongs to the desired landmarks, while other points should be discarded. This is done by filtering of the point cloud $P$ with adequate assumptions about the position of the landmarks, their size and the expected distribution of lidar points on the landmarks. 
\begin{figure}
    \centering
    \includegraphics[width=\linewidth]{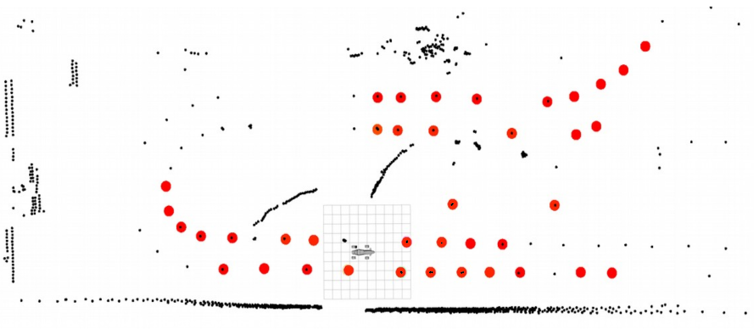}
    \caption{\small{Landmark detections: Black points show raw lidar points, red points are validated landmarks.}} \label{fig:detections}
\end{figure}
Analysis of the point density of $P$ showed that landmarks always occur in regions with a high point density. Therefore, a spatial clustering of all points was applied to extract features from the raw pointcloud (see \prettyref{fig:detections} for examples). We use the DBSCAN method \cite{MartinEster.1996} to extract features from $P$ according to their density. This leaves us with a set $C$ of $N_{\text{cluster}} \leq N_{\text{raw}}$ clusters of points. Each cluster $c_j$ is a subset of $P$, $c_j \subseteq P, \: \forall j\in{1,\ldots, N_{\text{cluster}}}$. This step also removes noise contained in $P$. For example, many ground detections are classified as noise, because of their sparse occurence. Each cluster $c_j$ includes information about the represented object. The position for each cluster is given by its centroid:
\begin{equation}
    \boldsymbol{\mu}_j = \boldsymbol{\mu}\left( c_j \right) = \frac{1}{\left| c_j \right |} \sum_{\boldsymbol{p} \in c_j} \boldsymbol{p}.
\end{equation}
All traffic cones have similar size. We utilize this fact by thresholding the cluster points variance. The empiric variance for one cluster is
\begin{equation}
    \mathbf{Var}\left(c_j\right) = \frac{1}{\left| c_j \right | - 1}\sum_{\boldsymbol{p} \in c_j}\left(\boldsymbol{p}-\boldsymbol{\mu}_j\right)^2.
\end{equation}
Setting maximal values for all directions removes big, high variance clusters. 

Another heuristic we employ considers the neighbourhood of each cluster. We set restrictions on the distance to the nearest neighbour of each cluster. This distance is defined as the distance between the centroids of the considered clusters. This can be done, because we already preselected clusters with approximately equal variance. We use the epsilon-neighbourhood \cite{MartinEster.1996} for each cluster,
\begin{equation}
    N\left(c_j, \varepsilon \right) = \left\lbrace c_k \in C \: | \: \left\Vert \boldsymbol{\mu}\left(c_j \right) - \boldsymbol{\mu}\left(c_k \right) \right\Vert^2 < \varepsilon  \right\rbrace
\end{equation}
as a measure and calculate $N\left(c_j, \varepsilon \right)$ for two different thresholds $\varepsilon_1>0$ and $\varepsilon_2 > \varepsilon_1$. The neighbourhood defined by $\varepsilon_1$ describes an inner region, where we allow other clusters. This is necessary to account for cases, where multiple clusters describe one object of interest. Above the threshold $\varepsilon_1$, we assume no other clusters, because usually the landmarks are separated by approximately $3.5 \, \text{m}$. So, $\varepsilon_2$ describes the (squared) clearance radius in which no other clusters should be present. This yields the following criterion for landmark clusters:
\begin{equation}
    \left| N\left(c_j, \varepsilon_1 \right)\right| \overset{!}{=} \left| N\left(c_j, \varepsilon_2 \right)\right|
\end{equation}
Applying all filters and neglecting the z-coordinate of the centroids $\boldsymbol{\mu}_i$ leaves a small set of potential landmarks, which are further treated as proposals.

\subsection{Landmark Validation}
All landmark proposals $\mathbf{p}_{l,i},\: i=1..N_\text{prop}$ are validated against features in the camera images. This is done by mapping the extracted landmark positions $\mathbf{p}_l=\left(x_l,y_l\right)^\transpose$ into pixel space, calculation of tight bounding boxes, and classification of these boxes.

\paragraph{Projection of Landmarks}
To compare landmark positions in the image plane, a projection $\boldsymbol{\phi}:\mathbb{R}^2 \rightarrow \mathbb{R}^2$, mapping from the ground plane to the 2D image plane has to be found. This is usually done by calculating the intrinsic and extrinsic parameters for each camera. We simplify this process by approximating $\boldsymbol\phi$ with a polynomial of degree $N$:
\begin{equation}
    \boldsymbol\phi \left( \mathbf{p}_l = (x_l,y_l)^\transpose \right) \approx \sum_{i=0}^N\sum_{j=0}^N \boldsymbol a_{i,j} x_l^iy_l^j .
\end{equation} 
This results in $2N^2$ parameters, that have to be tuned. We used a setup with multiple landmarks registered in both lidar and camera to regress these parameters $\boldsymbol a_{i,j}$. The projection maps landmark positions $\mathbf{p}_l$ to pixel coordinates $(u,v)^\transpose=\boldsymbol\phi \left(\mathbf{p}_l\right)$. We assume these coordinates to be the center of the landmarks.

\begin{figure}
    \begin{subfigure}[b]{0.49\linewidth}
         \centering
         \includegraphics[width=\linewidth]{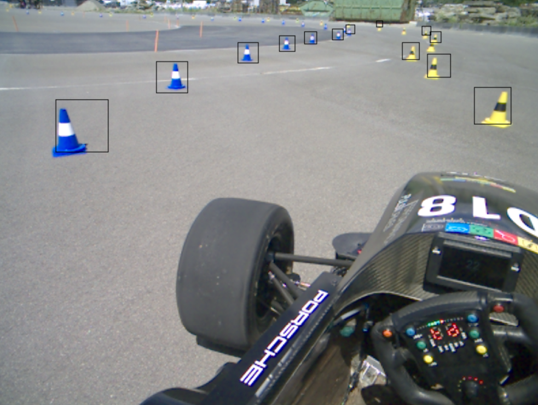}
         \caption{}
         \label{fig:bbs_a}
     \end{subfigure}
     \begin{subfigure}[b]{0.49\linewidth}
         \centering
         \includegraphics[width=\linewidth]{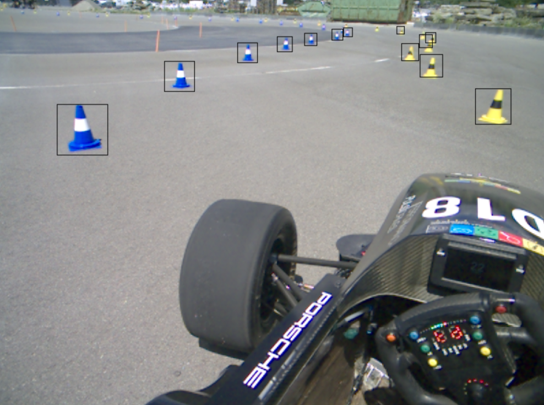}
         \caption{}
         \label{fig:bbs_b}
     \end{subfigure}
     \caption{\small{Landmark bounding boxes: (a) shows an example track with regressed bounding boxes. (b) shows the same image, but with corrected bounding boxes.}}
\end{figure}

\paragraph{Bounding Box Regression}
To achieve a proper validation of all landmarks, a boundary of them has to be calculated. A standard approach is the  bounding box, which creates a rectangular boundary around the object of interest. In addition to the already calculated center of the bounding box,, that is calculated by the projection method outlined above, a size estimate has to be made, to specify the boundary. Here we make use of the fact, that all objects should have the same aspect ratio, and that the scaling of each landmark inside the image is reciprocal to the distance of this landmark to the origin (vehicle rear axis). Furthermore, we assume an anisotropic scaling behavior and therefore introduce one scaling factor per direction, leading to the following approximations for width $w_l$ and height $h_l$ of bounding box $l$:
\begin{equation}
    w_l = \frac{s_u}{ \left\Vert \mathbf{p}_l \right\Vert}, \quad 
    h_l = \frac{s_v}{ \left\Vert \mathbf{p}_l \right\Vert},
\end{equation}
with scaling factors $s_u$ and $s_v$. We use the resulting bounding box $\left( u, v, w_l, h_l \right)$ as a rough estimate of the ideal boundary (see \prettyref{fig:bbs_a}). This works reasonably well but is not yet robust enough. To reduce the error introduced by the assumptions made above, we correct the bounding box with a simple, yet efficient method. We make use of the unique color of the landmarks, to apply an area approximation for each specific landmark. We calculate the centroid of this area and shift the bounding box accordingly. This significantly improves the bounding box estimation (see \prettyref{fig:bbs_b}).

\paragraph{Landmark classification}
The final step for validation is a classification of each bounding box. According to the rules \cite{fsg2019rules}, there are five possible cases:
\begin{enumerate}
    \item small blue traffic cone,
    \item small yellow traffic cone,
    \item small orange traffic cone,
    \item big orange traffic cone,
    \item none of them ( false positive proposal).
\end{enumerate}
This task is well established. We decided to use a Convolutional Neural Network (CNN), to create a robust classifier. Following our main goal of efficiency, we used an adapted MobileNet V2 \cite{Sandler.}, which was further optimized for our application. Training was done by using network-based transfer learning \cite{Oquab.2014}, with ImageNet \cite{Russakovsky.2014} as a source domain for all convolutional layers. The network is finetuned on a semi-manually annotated dataset.

\paragraph*{Dataset} Our dataset contains images that were taken with KIT19d's camera setup and also similar images from our previous driverless vehicle. Bounding boxes of landmarks are annotated automatically, using our bounding box regression setup. Each bounding box is further annotated with a quality label to indicate how difficult the classification would be for a human. We used a scale from $0$ (classification not possible) to 10 (easy, high resolution). False samples (label \textit{none}) were created by sampling false positive samples, proposals which consisted of random internet images with the same colors, and random crops of captured images.

\paragraph*{Optimization} We used a semi-factorial, manual parameter-search for all hyper-parameters. During training, we noticed a tendency to only rely on color information, which is an easy way for classification, but not reliable on low contrast image regions or against false positives with similar color (e.g. human legs). To guide our classifier, we preselected low-quality images with heavy augmentation to force more complex decision rules. This led to an increase in accuracy in all categories.

%% file: Sections/07_Localization_and_Mapping.tex
\section{Localization and Mapping}\label{sec:slam}
The localization and mapping algorithm tracks the position of all observed landmarks and provides a transformation between the body-fixed coordinate system of the car and a world-fixed coordinate system located at the beginning of the track. Based on this information, the motion planning algorithm calculates a suitable path through the landmarks. 

Assuming the pose of the vehicle is known, well-established methods exist for generating a map from observations of the environment \cite[\S 9]{thrun2005robotics}. The same applies for the opposite case, where a map of the environment is available and the pose of the vehicle is unknown \cite[\S 7]{thrun2005robotics}. However, a much more difficult problem arises when both the map and the vehicle pose are unknown. This problem, which is commonly referred to as Simultaneous Localization and Mapping (SLAM), is subject to ongoing research and a selection of methods to solve it is for example presented in \cite{thrun2005robotics}. 

Solving the SLAM problem is a computationally expensive task. Furthermore, deploying the algorithm in a race car requires higher update rates compared to other applications (e.g. indoor robots). However, our application is very confined and allows us to make several assumptions that we can leverage to create a simpler algorithm. In this section, an Extended Kalman Filter (EKF) for the pose estimation of the vehicle is proposed. In our case, the estimation is sufficiently accurate to generate a global map during the Autocross race without further correction of the landmark positions. Thus, in contrast to a SLAM algorithm, the landmark positions are not included in the state vector.

\subsection{Extended Kalman Filter}
The system uses an Extended Kalman Filter (EKF) to track the pose of the vehicle. We assume that the race track is flat, which is valid for most of the competitions sites, therefore the state vector contains the x- and y-coordinates and the yaw angle \(\psi\) of the vehicle.
\begin{equation}
\mathbf{x}_k = (x, y, \psi)^\transpose
\end{equation}

\paragraph{Prediction}
Instead of modeling the vehicle's response to an input from the autonomous system, the rear wheel speeds \(n_{rl}\) and \(n_{rr}\) and yaw rate \(\dot{\psi}\) are used directly in the prediction step.
The state transition is given by
\begin{equation}
    \mathbf{\hat{x}}_{k+1} = f(\mathbf{x}_k, \mathbf{u}_k) + \epsilon_k,
\end{equation}

with

\begin{equation}
    f(\mathbf{x}_k, \mathbf{u}_k) = \mathbf{x}_k + \mathbf{B}_k \mathbf{u}_k,
\end{equation}


where $\epsilon$ is the zero mean gaussian process noise and \(\mathbf{u}_k = (n_{rl}, n_{rr}, \dot{\psi})^\transpose\) denotes the input vector which is transformed into the state-space by 

\begin{equation}
    \mathbf{B}_k = \begin{pmatrix} 
    \pi r_{dyn} \cos \psi &  \pi r_{dyn} \cos \psi & 0 \\
    \pi r_{dyn}\sin \psi & \pi r_{dyn}\sin \psi & 0 \\
    0 &  0 & 1
    \end{pmatrix} \Delta t.
\end{equation}

Here \(r_{dyn}\) represents the dynamic radius of the wheel.

\paragraph{Correction} The predicted state vector $\mathbf{\hat{x}}_{k+1}$ is augmented with the position of each of the $N_{\text{map}}$ fixed landmarks $\mathbf{m}_{l,i}$ contained in the map $M$. The map is structured as follows:

\begin{equation}
    M=\left\lbrace \mathbf{m}_{l,i} = (x_{m,i}, y_{m,i})^\transpose \: | \: i=1,2,\ldots,N_{\text{map}}\right\rbrace.
\end{equation}

Equation (13) describes the resulting composition of the augmented state vector.

\begin{equation}
    \mathbf{\hat{x}}_{k+1}^{Aug} = (x, y, \psi, \mathbf{m}_{{l,1}}, \mathbf{m}_{{l,2}}, ... ,\mathbf{m}_{l,N_{\text{map}}})^\transpose.
\end{equation}

Furthermore, a landmark proposal $\mathbf{p}_{l}$ is only considered in the correction step if it was tracked $n > 1$ times in the same color. The observation model for a set of measurements $\mathbf{z}_k = (\mathbf{p}_{l,0}, ..., \mathbf{p}_{l,N_{\text{prop}}})^\transpose$ is given by

\begin{equation}
    \mathbf{z}_k = h(\mathbf{\hat{x}}_{k+1}) + \delta_k,
\end{equation}

with

\begin{equation}
h(\mathbf{\hat{x}}_{k+1})
= \begin{pmatrix}
\mathbf{R_\psi^{-1}} \cdot \begin{pmatrix} x_{m,1} - x \\ y_{m,1} - y \end{pmatrix} \\
\vdots \\
\mathbf{R_\psi^{-1}} \cdot \begin{pmatrix} x_{m,N_{\text{map}}} - x \\ y_{m,N_{\text{map}}} - y \end{pmatrix} \\
\end{pmatrix},
\end{equation}

The matrix $\mathbf{R_\psi^{-1}}$ describes a 2D rotation with respect to the yaw angle $\psi$ in the negative direction.


The observation noise is assumed to be a zero mean gaussian noise and denoted as $\delta_k$. Based on the difference between the measurement and the prediction, the state is then updated according to the EKF update step as described by Thrun et al. \cite{thrun2005robotics}. 

\paragraph{Data association}
The correction step relies on the knowledge of the correspondence between the mapped landmarks and the measured landmarks. However, finding the correct associations is challenging and false associations will cause the filter to diverge.
\\
A common approach to solve this problem is the  Individual Compatibility Nearest Neighbor (ICNN) algorithm \cite{neira2001jcbb}. However, according to Neira and Tardos in \cite{neira2001jcbb}, the pose estimate error must not be greater than the distance between features. Neither the FSG rules \cite{fsg2019rules} nor the competition handbook \cite{fsg2019handbook} specify a minimum distance between cones, but past competitions have shown that their spacing can be as low as \SI{30}{\centi\meter}. Especially in tight corners, the error of the pose estimate can approach this distance and thus make the ICNN algorithm susceptible to false associations. Therefore, the computationally more expensive, but also more robust Joint Compatibility Branch and Bound (JCBB) algorithm \cite{neira2001jcbb} was used. 

%% file: Sections/08_Motion_Planning_and_Control.tex
\section{Motion Planning and Control}\label{sec:motion_planning}

Motion planning is concerned with the problem of finding a geometric path and the velocity at which it should be traveled starting at an initial pose and ending in a goal region. It is additionally required for the path to not intersect with obstructed areas of the configuration space which corresponds to staying inside the boundaries of the race track for autonomous racing applications. Static and dynamic obstacles are often also considered part of the problem but are not relevant in the Formula Student context. While it is possible to formulate this problem in terms of the control inputs of the vehicle to merge motion planning and control into a single problem, it is common practice to treat these separately. Paden et al. \cite{paden2016survey} give an overview of state-of-the-art motion planning and control techniques. In both motion planning and control, optimization-based methods are well suited for racing applications. We consider geometric path optimization to be of minor importance due to the small performance benefit to be gained between the narrow boundaries of typical Formula Student circuits. In contrast, optimization-based vehicle control is a crucial component for the performance of our vehicle, since it allows for highly dynamic maneuvers. We have therefore decided to combine a fairly simple motion planning method with a model predictive control (MPC) algorithm for lateral vehicle control. For the sake of simplicity, longitudinal control is decoupled from the MPC problem. Since the longitudinal dynamics of a race car are represented well by a double integrator with a velocity-dependent offset, a feedforward PI controller is a suitable choice for tracking a predefined velocity target.

\subsection{Motion Planner}

We use a simple motion planner that provides a geometric reference path to be tracked by the lateral controller and also the corresponding velocity target as reference for longitudinal control. Following the path-velocity-decomposition method these two reference trajectories are determined sequentially. As we consider geometric path optimization to be of minor importance, the centerline of the track ahead of the vehicle is used as the reference path. The next step is to attribute a target velocity to this reference path that is as fast as possible but also ensures that the vehicle is operated safely inside its performance envelope. Lap time simulation methods lend themselves to solve this kind of problem and in particular quasi-steady-state methods are suitable for online applications. This is because a large fraction of the method can be computed offline. A sophisticated nonlinear vehicle model is used to find the combined lateral and longitudinal acceleration limits of the vehicle for different speeds which is often referred to as GGS-data. In essence this provides an acceleration limit map of the vehicle which is used by the online algorithm to find a velocity trajectory that obeys those limitations. We parameterize the method such that if the velocity target is tracked closely by the real vehicle, the tires do not enter operation regions where tire behavior is highly nonlinear to ensure that the linear vehicle model used in the lateral controller is valid. The remainder of this section describes the lateral controller.

\subsection{Predictive Steering Controller}
\begin{figure}
	\centering
	\includegraphics[width=1.0\columnwidth]{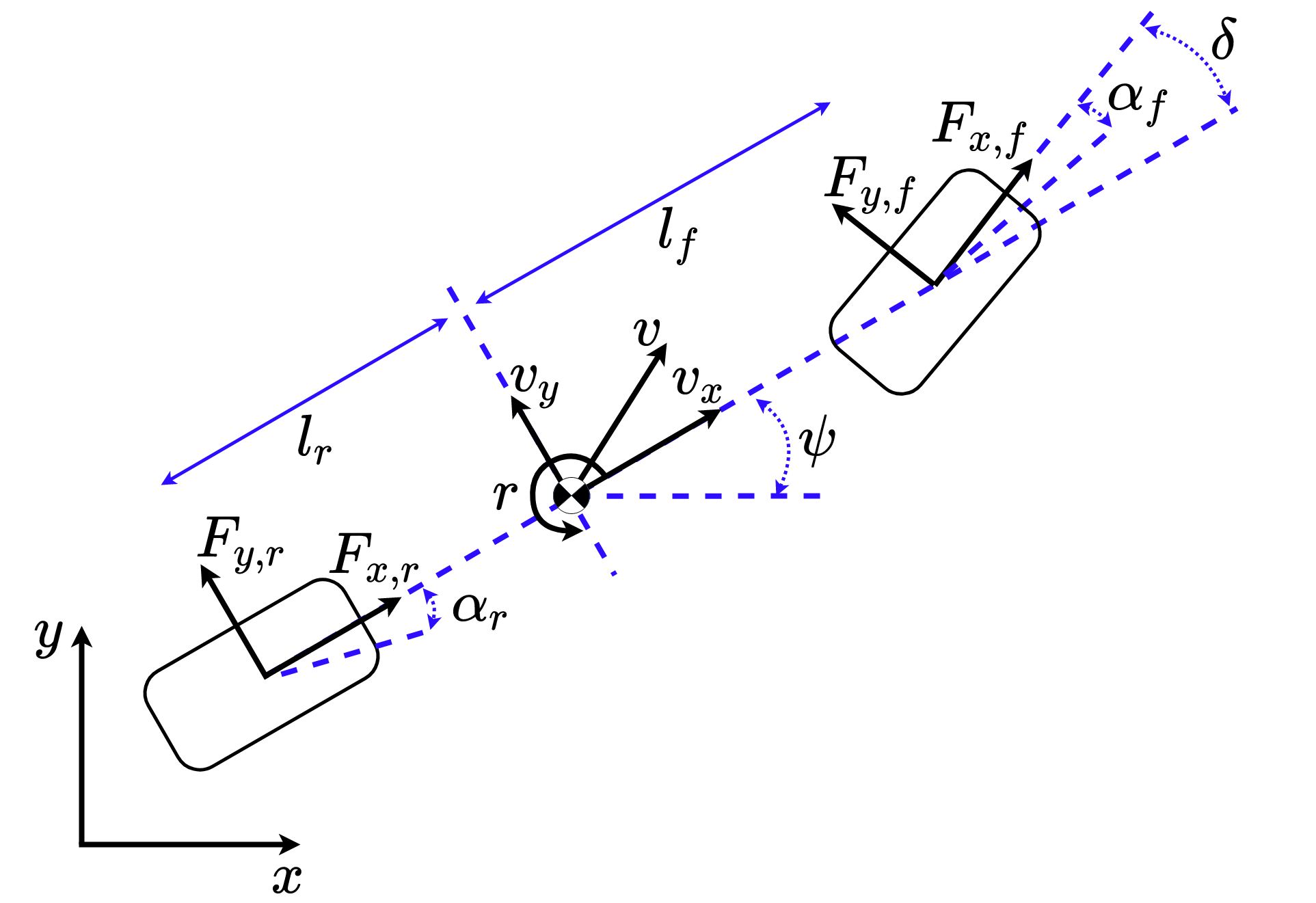}
	\caption{\small{Dynamic bicycle model used by the predictive steering controller.}}
	\label{fig:MPC_vehicle_model}
\end{figure}
Having used a Stanley controller \cite{hoffmann2007autonomous} in previous years that is based on the steering kinematic of the vehicle and uses a single point on the target path as reference, a model predictive steering controller was developed for KIT19d. This choice was motivated by the goal to make better use of the knowledge about the path ahead of the vehicle and the possibility to use more realistic vehicle models as the basis for the controller. We start the discussion about controller design with the internal model used for making predictions. Vehicle motion is simplified to exist in an SE(2) configuration space, which means the vehicle has two translational and one rotational degree of freedom. Thereby, body motions such as pitch, roll and heave are neglected and the surface on which the vehicle moves is assumed to be planar, which is usually a valid assumption for Formula Student tracks. Since we use MPC only as a steering controller, further simplification can be achieved by excluding the longitudinal vehicle dynamics from the model. However to associate each time step on the horizon with the desired path coordinate it is necessary to prescribe longitudinal velocity which can be retrieved from the velocity target used for the longitudinal controller. An even simpler and in our tests close to equivalent method is to use the current velocity measurement and keep it constant over the whole prediction horizon. Using the dynamic bicycle model illustrated in \prettyref{fig:MPC_vehicle_model}, the following set of nonlinear differential equations then describes the remaining lateral and yaw dynamics:

\begin{align}
\dot{y} &= v_x  \sin{\psi} + v_y \cos{\psi} \\
\dot{v}_y &= \frac{1}{m} (F_{y,f} \cos{\delta} + F_{y,r}) - v_x r \\
\dot{\psi} &= r \\
\dot{r} &= \frac{1}{I_z} M_z 
\end{align}

with lateral tire forces
\begin{equation}
    F_{y,f} = C_f \alpha_f, \quad F_{y,r} = C_r \alpha_r, \\
\end{equation}

yaw moment
\begin{equation}
    M_z = l_f F_{y,f} \cos{\delta} - l_r F_{y,r}, \\
\end{equation}

and tire slip angles
\begin{align}
    \alpha_f &= -\delta + \arctan{\left( \tfrac{v_y + l_f \dot{\psi}}{v_x} \right)}, \\ \alpha_r &= \arctan{\left( \tfrac{v_y - l_r \dot{\psi}}{v_x} \right)}.
\end{align}

With the state vector $\boldsymbol{x} = [y, v_y, \psi, r]^\transpose$ and the steering angle as scalar input $u = \delta$ the linearization of this system around straight line driving reads:

\begin{equation}
\dot{\boldsymbol{x}} = \mathbf{A} \boldsymbol{x} + \mathbf{b} u,
\label{eq:ode}
\end{equation}

with

\begin{equation}
\mathbf{A} = \begin{bmatrix}0 & 1 & v & 0 \\ 0 & \frac{C_f + C_r}{m v} & 0 & \frac{l_f C_f - l_r C_r}{m v}-v \\ 0 & 0 & 0 & 1 \\ 0 & \frac{l_f C_f - l_r C_r}{I_z v} & 0 & \frac{l_f^2 C_f + l_r^2 C_r}{I_z v}\end{bmatrix},
\end{equation}

\begin{equation}
\mathbf{b} = \begin{bmatrix}0 \\ \frac{-C_f}{m} \\ 0 \\ \frac{-l_f C_f}{I_z}\end{bmatrix}.
\end{equation}

This is a continuous-time model which has to be discretized due to the transcription of the MPC problem from a dynamic optimization problem to a parameter optimization problem. Discretization can be achieved by using numerical integrators such as the forward Euler scheme or the Runge-Kutta methods. However, for linear models it is also possible to obtain a discretization by using matrix exponentials that result from solving the differential equation \eqref{eq:ode} over a single time step using a piecewise constant representation of the input variables. 

We formulate and solve a linear time-varying MPC (LTV-MPC) problem, largely following the methods presented in \cite{maciejowski2002predictive}. Using the time-discrete linear vehicle model, the receding horizon optimal control problem can directly be stated as a parameter optimization problem. It is formulated in terms of the control input rate $\Delta u$ from which $u$ is obtained by accumulation: 

\begin{subequations}
	\label{eq:LTV_MPC}
	\begin{alignat}{2}
	& \underset{\substack{\Delta u_{1:N}, \\ \boldsymbol{x}_{1:N+1}}}{\text{min}} \
	& & \sum^{N}_{k=1} \left(\Vert \boldsymbol{x}_k-\boldsymbol{x}^{ref}_k \Vert^2_\mathbf{Q} \nonumber + R \Delta u_k^2 \right)   \\
	\label{eq:LTV_MPCa}
	&
	& & \qquad \qquad \qquad + \Vert \boldsymbol{x}_{N+1}-\boldsymbol{x}^{ref}_{N+1} \Vert^2_\mathbf{P} \\
	\label{eq:LTV_MPCb}
	& \text{s.t.}
	& & \boldsymbol{x}_{k+1} = \mathbf{A} \boldsymbol{x}_k + \mathbf{b} u_k \quad k=1,..,N\\
	\label{eq:LTV_MPCc}
	&
	& & u_{k+1} = u_k + \Delta u_k \quad k=1,..,N\\
	\label{eq:LTV_MPCd}
	&
	& & \boldsymbol{x}_1 = \hat{\boldsymbol{x}}\\
	\label{eq:LTV_MPCe} 
	&
	& & u_1 = \hat{u}\\
	\label{eq:LTV_MPCf}
	&
	& & \mathbf{D} \boldsymbol{x}_k + \mathbf{e} u_k + \mathbf{f} \leq \mathbf{0} \quad k=1,..,N+1\\
	\label{eq:LTV_MPCg}
	&
	& & \underline{\boldsymbol{x}} \leq \boldsymbol{x}_k \leq \overline{\boldsymbol{x}} \quad k=1,..,N+1\\
	\label{eq:LTV_MPCh}
	&
	& & \underline{u} \leq u_k \leq \overline{u} \quad k=1,..,N+1 \\
	\label{eq:LTV_MPCi}
	&
	& & \underline{\Delta u} \leq \Delta u_k \leq \overline{\Delta u} \quad k=1,..,N
	\end{alignat}
\end{subequations}

This formulation follows the direct simultaneous discretization approach \cite[\S 5.4.2]{chachuat2007nonlinear}, since it includes the state variables in the set of optimization variables and enforces the system dynamics explicitly by equality constraints \eqref{eq:LTV_MPCb}. In such formulations the Hessian of the objective function and the constraint matrices are block diagonal if the variables are sorted appropriately. Quadratic programming (QP) solvers that exploit this structure can therefore solve the problem efficiently (e.g. OSQP \cite{stellato2018osqp}, HPIPM \cite{frison2020hpipm}, qpDUNES \cite{frasch2015parallel}, FORCES \cite{domahidi2012efficient}). We have however opted for an approach that does not require a sophisticated QP solver. Maciejowski \cite[\S 2.6]{maciejowski2002predictive} presents how the state variables can be eliminated from the problem by using the linear system dynamics to express them in terms of the control variables. In contrast to formulation \eqref{eq:LTV_MPC} where the system dynamics are only satisfied upon convergence, this leads to the dynamics being always satisfied even before the optimal solution is found. Thus, this process yields a formulation following the direct sequential approach \cite[\S 5.4.3]{chachuat2007nonlinear}. The elimination of the state variables transforms the originally sparse problem into a smaller problem with dense matrices and is therefore also known as condensing. The resulting problem has the following form:

\begin{subequations}
	\label{eq:denseQP}
	\begin{alignat}{2}
	\label{eq:denseQPa}
	& \underset{\boldsymbol{\Delta u}_{1:N}}{\text{min}} \quad
	& &  \frac{1}{2} \boldsymbol{\Delta u}^\transpose_{1:N} \mathbf{H} \boldsymbol{\Delta u}_{1:N} + \mathbf{g}^\transpose \boldsymbol{\Delta u}_{1:N}\\
	\label{eq:denseQPb}
	& \text{s.t.} \quad
	& & \underline{\mathbf{d}} \leq \mathbf{D}\boldsymbol{\Delta u}_{1:N} \leq \overline{\mathbf{d}} \\
	\label{eq:denseQPc}
	& 
	& & \underline{\Delta u} \leq \Delta u_k \leq \overline{\Delta u}
	\end{alignat}
\end{subequations}

Here all decision variables have been gathered in a single vector $\boldsymbol{\Delta u}_{1:N} = [\Delta u_1, \Delta u_2, ..., \Delta u_N]^\transpose$. Problem \eqref{eq:denseQP} is a dense quadratic program and can be solved by general-purpose QP solvers such as qpOASES \cite{ferreau2014qpoases}. It is desirable for this QP to be strictly convex as in this case the problem is guaranteed to have a unique global minimum. Additionally the well-known KKT-conditions provide a sufficient condition of optimality for convex problems. As shown in \cite[p.76]{maciejowski2002predictive}, problem \eqref{eq:denseQP} is strictly convex if the input weight matrix is positive definite which translates into $R>0$ for our single-input case. Constraints on the state and control variables are encoded in the linear inequalities \eqref{eq:denseQPb}. For our application these could be used to implement lower and upper bounds on the steering angle. Experience from previous vehicles and simulation results on circuits representative for Formula Student competitions do however suggest that this limit is hardly ever reached when operating the vehicle within its handling limits. The constraints of problem \eqref{eq:denseQP} were therefore dropped, leading to an unconstrained QP for which the solution is obtained by solving the following set of linear equations:

\begin{equation}
    \mathbf{H} \boldsymbol{\Delta u}_{1:N} = -\mathbf{g}.
    \label{eq:linearSystem}
\end{equation}

There exist several direct and iterative methods for solving such linear systems. We have chosen to implement the method with the Eigen3 C++ library \cite{eigenweb} and use its Cholesky decomposition algorithm to solve problem \eqref{eq:linearSystem}.

This basic MPC controller can be enhanced by a delay compensation which is often used to make up for the time needed to solve the optimization problem. In our case of solving an unconstrained QP the delay introduced by optimization is not significant but the same techniques can be used to compensate for delays in between the controller and the actual steering angle. We have employed the approach from \cite[\S 2.5]{maciejowski2002predictive} where the state vector of the plant model is extended by a transport chain of input samples. A general real numbered delay $t_D$ is separated into an integer multiple of the discetization step size $T$ and a real numbered residual leading to $\lfloor \tfrac{t_D}{T} \rfloor +1$ new state vector entries.

Several methods that guarantee stability of MPC controllers for setpoint stabilization are well-established. Most of them are derived from the properties of infinite-horizon controllers for which from Bellman's principle of optimality, stability and recursive feasibility can be infered, if the initial optimization problem can be solved \cite[\S 6.2]{maciejowski2002predictive}. In receding-horizon control, penalties and constraints on the terminal state are used to obtain these characteristics. Also sufficiently long horizon lengths can be used to derive the required guarantees. However, since the correct choice of the terminal set, the weights of terminal penalties or the minimum horizon length is a nontrivial task, rigorous proofs of stability are often omitted in practice. Moreover, these proofs are often invalidated in the presence of disturbances and plant-model mismatch.  This is especially true when linear MPC is used to control highly nonlinear systems. Strictly speaking stability proofs for setpoint stabilization are not even applicable for our use case of trajectory tracking, for which theoretical analysis can be found in \cite{faulwasser2012optimization}. We have therefore chosen a more heuristic approach and focused on setting the parameters of the velocity planner such that the vehicle's dynamics are close to linear to keep the plant-model mismatch small. A sufficiently long horizon of $N=65$ with a sampling time of $\Delta t=\SI{20}{\milli\second}$ has been used. While terminal region constraints are not possible for our unconstrained formulation, terminal penalties can be included. Extensive simulation studies to verify stability and robustness of the approach were conducted before testing on the real vehicle.

%% file: Sections/09_Results.tex
\section{Results}\label{sec:results}

\begin{figure}
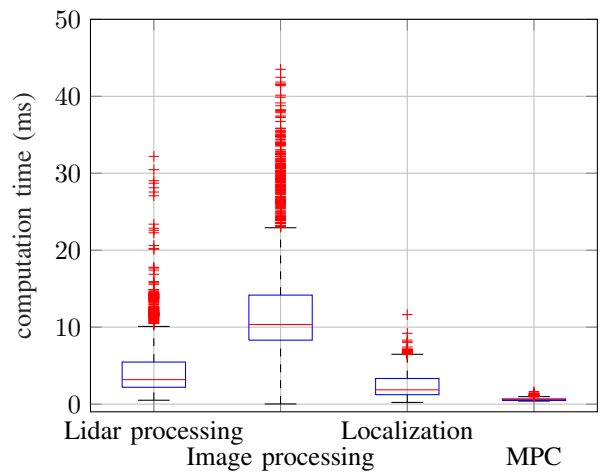

    \centering
    \setlength\fwidth{0.8\columnwidth}
    \includestandalone{Graphics/timingBoxPlot}
    \caption{\small{Timing results for the AS modules presented in this paper.}}
    \label{fig:timingBoxPlot}
\end{figure}

To demonstrate the capabilities of our implementation, we decided to present two important aspects: the real-time feasibility of the presented algorithms and the performance of the final prototype in competitions. 
\prettyref{fig:timingBoxPlot} shows the runtime distributions for the subsystems presented in this paper that were collected on the ACU under real racing conditions. 

All systems achieve cycle times below \SI{50}{\milli\second} which ensures that no significant delays build up between perception and control. It can also be seen that the perception module is the most time-consuming component, as the median aggregated inference time is about \SI{13.5}{\milli\second}. The runtimes of about \SI{3.2}{\milli\second} for the lidar clustering and \SI{10.3}{\milli\second} for the image processing do not limit the overall performance, as their inference times are lower than the lidar's and camera's update rates respectively. These results show the efficiency of the presented perception pipeline compared to other algorithms commonly used in a Formula Student context, such as \cite{yolo9000}. Furthermore, the computation times of the localization algorithm and the controller are considerably shorter compared to the perception modules.

\begin{figure}
    \centering
    \setlength\fwidth{0.9\columnwidth}
    \includestandalone{Graphics/ggtest}
    \caption{\small{Accelerations during the FSG19 Autocross w/o prior knowledge.}}
    \label{fig:gg_diagrams}
\end{figure}

\begin{figure}
  \centering
  \includegraphics[width=.49\textwidth]{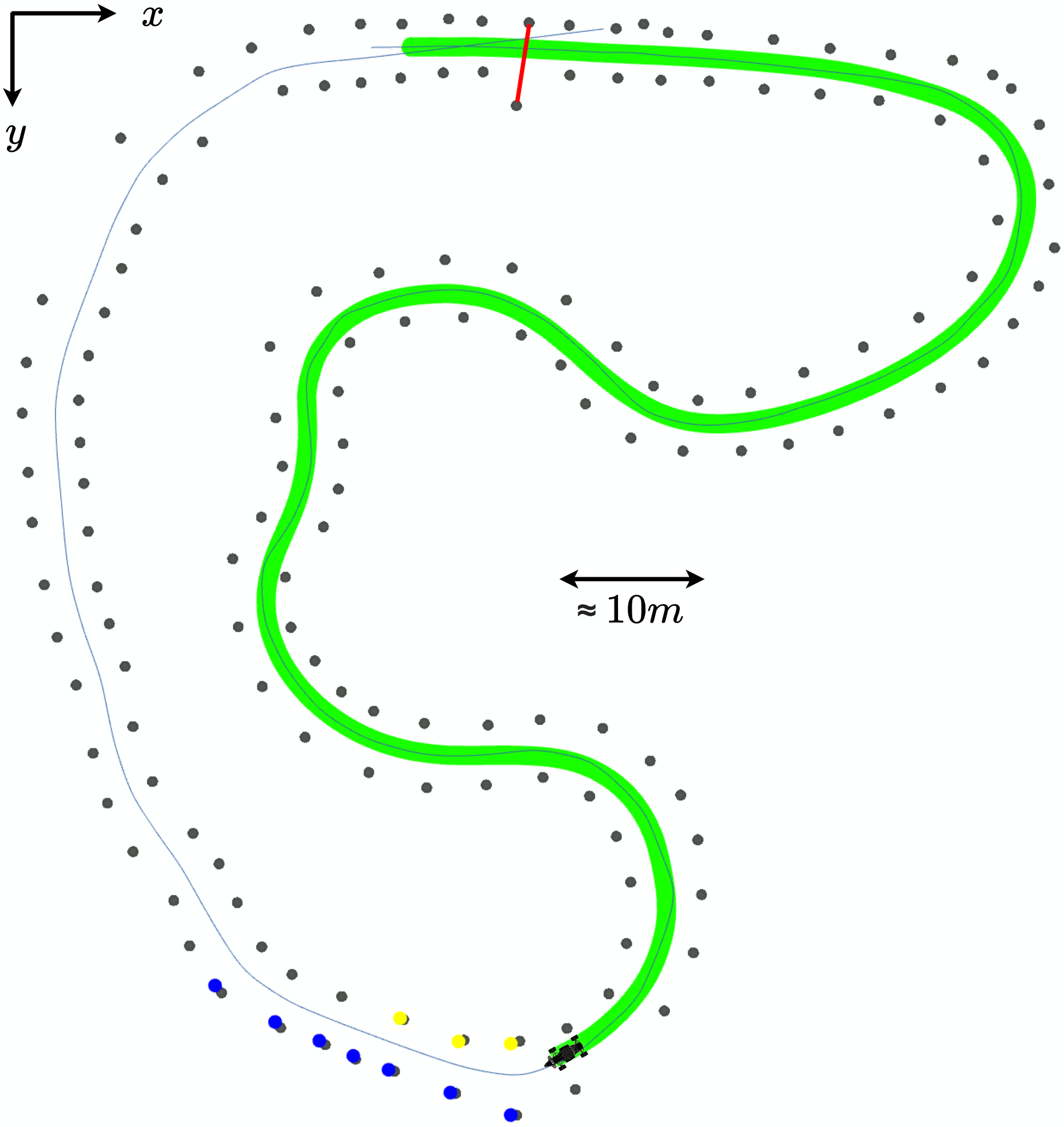}
  \caption{\small{Visualization of the KIT19d's first Autocross run at FSG19.}}
  \label{fig:autocross_mapping}
\end{figure}

\begin{table*}
\centering
\begin{tabular}{p{0.2\linewidth} | p{0.16\linewidth} p{0.16\linewidth} p{0.16\linewidth} p{0.16\linewidth} }
  \hline
    & Skidpad & Acceleration & Autocross \textbf{w/o} prior knowledge & Trackdrive \\
  \hline 
  \hline
  Zurich ETH & 5.992 + 2.00 & \textcolor{OliveGreen}{\textbf{3.597}} & 31.158 & 226.31 + 32.00   \\
  \textbf{Karlsruhe KIT} & \textcolor{OliveGreen}{\textbf{6.671 + 0.20}} & DNF & \textcolor{OliveGreen}{\textbf{28.186}} & \textcolor{OliveGreen}{\textbf{244.90 + 4.00}}   \\
  Delft TU & DNF & DNF & DNF & DNF   \\
  Augsburg UAS & DNF & 4.056 & DNF & 315.96 + 2.00   \\
  Hamburg TU & 8.993 + 0.20 & 18.322 & 65.356 + 6.00 & DNF   \\
  \hline
\end{tabular}
\caption{\small{Top 5 FSD Teams - Dynamic disciplines - Best lap times incl. penalties at FSG19.}}
\label{tab:results}
\end{table*}

Besides the latencies between perception and control, the performance of an autonomous racing vehicle on an unknown circuit is to a large extend determined by the maximum preview distance its perception system can provide. The presented algorithms yield a preview of up to \SI{42}{\meter}, which amounts to an increase of \SI{210}{\percent} compared to KIT19d's predecessor. To showcase the on-track performance of the vehicle, \prettyref{fig:gg_diagrams} presents logging data of the lateral and longitudinal accelerations and the velocity attained by the vehicle while driving the FSG19 Autocross without prior knowledge. Maximum lateral acceleration values come close to \SI{10}{\meter\per\second\squared} which is not far off the performance of a non-professional human driver. For comparison, when using a map, the vehicle achieved lateral accelerations only slightly over \SI{11}{\meter\per\second\squared} on the same circuit. \prettyref{fig:autocross_mapping} further illustrates the advantages of a high range perception system, as the vehicle had collected enough data to calculate the entire trajectory after completing less than \SI{70}{\percent} of the Autocross track.



    

The KIT19d won three out of four dynamic disciplines at the FSD 2019 competition in Hockenheim (see \prettyref{tab:results}) and achieved an overall first place at Formula Student Spain (FSS) in Barcelona. Compared to the competition season of 2018, the lap time\footnote{In 2018 the first run of the Trackdrive without prior knowledge was equivalent to the Autocross in 2019.} without prior knowledge was more than four times lower in 2019, even though the same hardware setup was used. Furthermore, the FSG19 Autocross time of the KIT19d, without prior knowledge, was 9\% faster than the second-fastest car. We would like to note that 9\% in terms of lap time is still a large margin.

%% file: Graphics/timingBoxPlot.tex
%
%
\begin{tikzpicture}

\begin{axis}[%
width=0.951\fwidth,
height=0.737\fwidth,
at={(0\fwidth,0\fwidth)},
scale only axis,
xmin=0.5,
xmax=4.5,
xtick={1,2,3,4},
xticklabels={{Lidar processing},{Image processing},{Localization},{MPC}},
x tick label style={yshift={-mod(\ticknum,2)*1em}},
ymin=-1,
ymax=50,
ylabel style={font=\color{white!15!black}},
ylabel={computation time (ms)},
axis background/.style={fill=white},
xmajorgrids,
ymajorgrids
]
\addplot [color=black, dashed, forget plot]
  table[row sep=crcr]{%
1	5.463\\
1	10.093\\
};
\addplot [color=black, dashed, forget plot]
  table[row sep=crcr]{%
2	14.16275\\
2	22.917\\
};
\addplot [color=black, dashed, forget plot]
  table[row sep=crcr]{%
3	3.32825\\
3	6.475\\
};
\addplot [color=black, dashed, forget plot]
  table[row sep=crcr]{%
4	0.67071\\
4	0.970118\\
};
\addplot [color=black, dashed, forget plot]
  table[row sep=crcr]{%
1	0.504\\
1	2.188\\
};
\addplot [color=black, dashed, forget plot]
  table[row sep=crcr]{%
2	0.018\\
2	8.31025\\
};
\addplot [color=black, dashed, forget plot]
  table[row sep=crcr]{%
3	0.217\\
3	1.22575\\
};
\addplot [color=black, dashed, forget plot]
  table[row sep=crcr]{%
4	0.395044\\
4	0.47036\\
};
\addplot [color=black, forget plot]
  table[row sep=crcr]{%
0.875	10.093\\
1.125	10.093\\
};
\addplot [color=black, forget plot]
  table[row sep=crcr]{%
1.875	22.917\\
2.125	22.917\\
};
\addplot [color=black, forget plot]
  table[row sep=crcr]{%
2.875	6.475\\
3.125	6.475\\
};
\addplot [color=black, forget plot]
  table[row sep=crcr]{%
3.875	0.970118\\
4.125	0.970118\\
};
\addplot [color=black, forget plot]
  table[row sep=crcr]{%
0.875	0.504\\
1.125	0.504\\
};
\addplot [color=black, forget plot]
  table[row sep=crcr]{%
1.875	0.018\\
2.125	0.018\\
};
\addplot [color=black, forget plot]
  table[row sep=crcr]{%
2.875	0.217\\
3.125	0.217\\
};
\addplot [color=black, forget plot]
  table[row sep=crcr]{%
3.875	0.395044\\
4.125	0.395044\\
};
\addplot [color=blue, forget plot]
  table[row sep=crcr]{%
0.75	2.188\\
0.75	5.463\\
1.25	5.463\\
1.25	2.188\\
0.75	2.188\\
};
\addplot [color=blue, forget plot]
  table[row sep=crcr]{%
1.75	8.31025\\
1.75	14.16275\\
2.25	14.16275\\
2.25	8.31025\\
1.75	8.31025\\
};
\addplot [color=blue, forget plot]
  table[row sep=crcr]{%
2.75	1.22575\\
2.75	3.32825\\
3.25	3.32825\\
3.25	1.22575\\
2.75	1.22575\\
};
\addplot [color=blue, forget plot]
  table[row sep=crcr]{%
3.75	0.47036\\
3.75	0.67071\\
4.25	0.67071\\
4.25	0.47036\\
3.75	0.47036\\
};
\addplot [color=red, forget plot]
  table[row sep=crcr]{%
0.75	3.183\\
1.25	3.183\\
};
\addplot [color=red, forget plot]
  table[row sep=crcr]{%
1.75	10.344\\
2.25	10.344\\
};
\addplot [color=red, forget plot]
  table[row sep=crcr]{%
2.75	1.859\\
3.25	1.859\\
};
\addplot [color=red, forget plot]
  table[row sep=crcr]{%
3.75	0.59176\\
4.25	0.59176\\
};
\addplot [color=black, draw=none, mark=+, mark options={solid, red}, forget plot]
  table[row sep=crcr]{%
1	10.478\\
1	10.5\\
1	10.5\\
1	10.534\\
1	10.536\\
1	10.639\\
1	10.761\\
1	10.774\\
1	10.817\\
1	10.875\\
1	10.883\\
1	10.921\\
1	10.948\\
1	10.987\\
1	11.077\\
1	11.095\\
1	11.109\\
1	11.16\\
1	11.21\\
1	11.237\\
1	11.27\\
1	11.271\\
1	11.316\\
1	11.321\\
1	11.334\\
1	11.349\\
1	11.361\\
1	11.363\\
1	11.392\\
1	11.425\\
1	11.438\\
1	11.44\\
1	11.443\\
1	11.452\\
1	11.453\\
1	11.502\\
1	11.509\\
1	11.527\\
1	11.542\\
1	11.558\\
1	11.563\\
1	11.567\\
1	11.572\\
1	11.573\\
1	11.583\\
1	11.591\\
1	11.598\\
1	11.618\\
1	11.62\\
1	11.633\\
1	11.637\\
1	11.653\\
1	11.659\\
1	11.664\\
1	11.686\\
1	11.687\\
1	11.69\\
1	11.694\\
1	11.694\\
1	11.711\\
1	11.715\\
1	11.721\\
1	11.725\\
1	11.729\\
1	11.731\\
1	11.743\\
1	11.751\\
1	11.756\\
1	11.767\\
1	11.773\\
1	11.774\\
1	11.79\\
1	11.793\\
1	11.793\\
1	11.797\\
1	11.815\\
1	11.817\\
1	11.82\\
1	11.829\\
1	11.837\\
1	11.838\\
1	11.84\\
1	11.842\\
1	11.844\\
1	11.847\\
1	11.85\\
1	11.853\\
1	11.856\\
1	11.857\\
1	11.893\\
1	11.897\\
1	11.917\\
1	11.924\\
1	11.927\\
1	11.957\\
1	11.967\\
1	12.01\\
1	12.011\\
1	12.022\\
1	12.049\\
1	12.068\\
1	12.078\\
1	12.079\\
1	12.08\\
1	12.133\\
1	12.141\\
1	12.145\\
1	12.167\\
1	12.178\\
1	12.182\\
1	12.23\\
1	12.236\\
1	12.256\\
1	12.281\\
1	12.283\\
1	12.352\\
1	12.39\\
1	12.467\\
1	12.469\\
1	12.474\\
1	12.475\\
1	12.491\\
1	12.541\\
1	12.568\\
1	12.585\\
1	12.59\\
1	12.595\\
1	12.619\\
1	12.62\\
1	12.757\\
1	12.965\\
1	12.989\\
1	13.004\\
1	13.167\\
1	13.245\\
1	13.317\\
1	13.355\\
1	13.439\\
1	13.455\\
1	13.498\\
1	13.537\\
1	13.599\\
1	13.608\\
1	13.631\\
1	13.665\\
1	13.694\\
1	13.736\\
1	13.75\\
1	13.758\\
1	13.78\\
1	13.782\\
1	13.802\\
1	13.809\\
1	13.825\\
1	13.837\\
1	13.88\\
1	13.882\\
1	13.885\\
1	13.893\\
1	13.915\\
1	13.918\\
1	13.968\\
1	13.969\\
1	13.981\\
1	13.985\\
1	14.012\\
1	14.046\\
1	14.064\\
1	14.064\\
1	14.085\\
1	14.122\\
1	14.164\\
1	14.177\\
1	14.187\\
1	14.19\\
1	14.214\\
1	14.238\\
1	14.257\\
1	14.266\\
1	14.277\\
1	14.283\\
1	14.331\\
1	14.35\\
1	14.371\\
1	14.381\\
1	14.392\\
1	14.45\\
1	14.509\\
1	14.611\\
1	14.834\\
1	14.845\\
1	14.872\\
1	15.565\\
1	15.739\\
1	15.864\\
1	15.904\\
1	16.87\\
1	17.416\\
1	17.677\\
1	17.779\\
1	20.13\\
1	20.201\\
1	20.602\\
1	22.262\\
1	22.584\\
1	22.845\\
1	23.381\\
1	27.065\\
1	27.556\\
1	28.111\\
1	28.703\\
1	29.029\\
1	30.463\\
1	32.188\\
};
\addplot [color=black, draw=none, mark=+, mark options={solid, red}, forget plot]
  table[row sep=crcr]{%
2	23\\
2	23.088\\
2	23.128\\
2	23.16\\
2	23.167\\
2	23.444\\
2	23.478\\
2	23.495\\
2	23.558\\
2	23.568\\
2	23.597\\
2	23.633\\
2	23.881\\
2	23.883\\
2	23.947\\
2	23.958\\
2	24.033\\
2	24.079\\
2	24.111\\
2	24.153\\
2	24.198\\
2	24.259\\
2	24.298\\
2	24.379\\
2	24.406\\
2	24.414\\
2	24.489\\
2	24.678\\
2	24.753\\
2	24.864\\
2	24.985\\
2	25.056\\
2	25.21\\
2	25.451\\
2	25.498\\
2	25.503\\
2	25.516\\
2	25.704\\
2	25.748\\
2	25.79\\
2	25.796\\
2	25.971\\
2	25.972\\
2	26.082\\
2	26.097\\
2	26.126\\
2	26.13\\
2	26.142\\
2	26.153\\
2	26.158\\
2	26.175\\
2	26.215\\
2	26.237\\
2	26.273\\
2	26.301\\
2	26.347\\
2	26.37\\
2	26.426\\
2	26.502\\
2	26.533\\
2	26.545\\
2	26.56\\
2	26.563\\
2	26.57\\
2	26.588\\
2	26.626\\
2	26.626\\
2	26.683\\
2	26.861\\
2	26.918\\
2	26.948\\
2	26.952\\
2	26.976\\
2	27.224\\
2	27.334\\
2	27.448\\
2	27.457\\
2	27.471\\
2	27.56\\
2	27.653\\
2	27.711\\
2	27.754\\
2	27.832\\
2	27.854\\
2	27.863\\
2	27.959\\
2	27.982\\
2	28.125\\
2	28.183\\
2	28.213\\
2	28.302\\
2	28.316\\
2	28.339\\
2	28.37\\
2	28.375\\
2	28.417\\
2	28.45\\
2	28.518\\
2	28.594\\
2	28.739\\
2	28.808\\
2	28.828\\
2	28.892\\
2	28.929\\
2	29.025\\
2	29.073\\
2	29.192\\
2	29.217\\
2	29.244\\
2	29.285\\
2	29.307\\
2	29.362\\
2	29.383\\
2	29.457\\
2	29.537\\
2	29.63\\
2	29.645\\
2	29.678\\
2	29.697\\
2	29.725\\
2	29.741\\
2	29.801\\
2	29.921\\
2	29.974\\
2	30.122\\
2	30.203\\
2	30.207\\
2	30.208\\
2	30.347\\
2	30.361\\
2	30.561\\
2	30.589\\
2	30.673\\
2	30.693\\
2	30.723\\
2	30.795\\
2	30.853\\
2	30.903\\
2	30.948\\
2	30.987\\
2	31.018\\
2	31.041\\
2	31.121\\
2	31.167\\
2	31.267\\
2	31.476\\
2	31.499\\
2	31.5\\
2	31.604\\
2	31.699\\
2	31.736\\
2	31.846\\
2	31.91\\
2	32.136\\
2	32.165\\
2	32.19\\
2	32.406\\
2	32.412\\
2	32.42\\
2	32.561\\
2	32.574\\
2	32.671\\
2	32.854\\
2	32.858\\
2	33.058\\
2	33.523\\
2	33.616\\
2	33.745\\
2	33.784\\
2	33.819\\
2	33.891\\
2	33.951\\
2	34.017\\
2	34.181\\
2	34.378\\
2	34.591\\
2	34.726\\
2	34.761\\
2	34.767\\
2	34.848\\
2	34.89\\
2	35.061\\
2	35.073\\
2	35.318\\
2	35.399\\
2	35.475\\
2	35.577\\
2	35.815\\
2	35.859\\
2	36.728\\
2	37.245\\
2	37.286\\
2	37.441\\
2	37.866\\
2	38.046\\
2	38.138\\
2	38.263\\
2	38.77\\
2	39.138\\
2	39.873\\
2	40.145\\
2	41.383\\
2	41.542\\
2	41.6\\
2	41.848\\
2	42.464\\
2	43.501\\
2	175.632\\
};
\addplot [color=black, draw=none, mark=+, mark options={solid, red}, forget plot]
  table[row sep=crcr]{%
3	6.486\\
3	6.492\\
3	6.501\\
3	6.502\\
3	6.507\\
3	6.509\\
3	6.517\\
3	6.529\\
3	6.537\\
3	6.556\\
3	6.559\\
3	6.561\\
3	6.564\\
3	6.564\\
3	6.564\\
3	6.574\\
3	6.578\\
3	6.582\\
3	6.585\\
3	6.602\\
3	6.606\\
3	6.606\\
3	6.612\\
3	6.613\\
3	6.621\\
3	6.623\\
3	6.625\\
3	6.627\\
3	6.648\\
3	6.65\\
3	6.654\\
3	6.657\\
3	6.657\\
3	6.658\\
3	6.672\\
3	6.672\\
3	6.675\\
3	6.677\\
3	6.679\\
3	6.68\\
3	6.684\\
3	6.685\\
3	6.692\\
3	6.698\\
3	6.699\\
3	6.7\\
3	6.712\\
3	6.714\\
3	6.716\\
3	6.718\\
3	6.72\\
3	6.722\\
3	6.726\\
3	6.739\\
3	6.742\\
3	6.742\\
3	6.749\\
3	6.75\\
3	6.75\\
3	6.755\\
3	6.76\\
3	6.761\\
3	6.761\\
3	6.766\\
3	6.783\\
3	6.785\\
3	6.79\\
3	6.79\\
3	6.795\\
3	6.799\\
3	6.818\\
3	6.824\\
3	6.832\\
3	6.838\\
3	6.845\\
3	6.857\\
3	6.867\\
3	6.869\\
3	6.874\\
3	6.889\\
3	6.896\\
3	6.909\\
3	6.911\\
3	6.911\\
3	6.929\\
3	6.93\\
3	6.938\\
3	6.975\\
3	7.013\\
3	7.015\\
3	7.023\\
3	7.032\\
3	7.057\\
3	7.067\\
3	7.084\\
3	7.123\\
3	7.126\\
3	7.136\\
3	7.416\\
3	8.031\\
3	8.088\\
3	8.318\\
3	9.196\\
3	11.632\\
};
\addplot [color=black, draw=none, mark=+, mark options={solid, red}, forget plot]
  table[row sep=crcr]{%
4	0.973441\\
4	0.975645\\
4	0.979053\\
4	0.979581\\
4	0.982406\\
4	0.982479\\
4	0.983066\\
4	0.984847\\
4	0.985007\\
4	0.985877\\
4	0.986395\\
4	0.988074\\
4	0.989708\\
4	0.990685\\
4	0.992008\\
4	0.9923\\
4	0.992485\\
4	0.993788\\
4	0.994097\\
4	0.997522\\
4	0.998332\\
4	0.998363\\
4	0.998898\\
4	0.999383\\
4	0.999413\\
4	0.99992\\
4	1.00049\\
4	1.00102\\
4	1.00391\\
4	1.00749\\
4	1.00765\\
4	1.01251\\
4	1.01428\\
4	1.0143\\
4	1.01471\\
4	1.01862\\
4	1.0241\\
4	1.02421\\
4	1.02495\\
4	1.02598\\
4	1.02901\\
4	1.03263\\
4	1.03312\\
4	1.03332\\
4	1.03395\\
4	1.03437\\
4	1.03702\\
4	1.03841\\
4	1.04136\\
4	1.04162\\
4	1.04276\\
4	1.04489\\
4	1.04536\\
4	1.04819\\
4	1.05016\\
4	1.05091\\
4	1.05773\\
4	1.05787\\
4	1.0603\\
4	1.06143\\
4	1.06871\\
4	1.06947\\
4	1.07044\\
4	1.07249\\
4	1.0736\\
4	1.07391\\
4	1.07473\\
4	1.07625\\
4	1.084\\
4	1.08441\\
4	1.0852\\
4	1.08991\\
4	1.09697\\
4	1.09839\\
4	1.09855\\
4	1.10068\\
4	1.10189\\
4	1.10223\\
4	1.1112\\
4	1.12779\\
4	1.12922\\
4	1.13292\\
4	1.13691\\
4	1.13835\\
4	1.14292\\
4	1.15511\\
4	1.15839\\
4	1.19009\\
4	1.24438\\
4	1.28413\\
4	1.30786\\
4	1.38055\\
4	1.41895\\
4	1.44355\\
4	1.63627\\
};
\end{axis}
\end{tikzpicture}%

%% file: Sections/10_Conclusion.tex
\section{Conclusion}\label{sec:conclusion}

We have presented the methods and concepts that were essential to the development of the software stack of the autonomous racing prototype KIT19d. Central algorithmic modules from perception, localization and control have been covered in depth. Selected performance evaluations of the final system and competition results have been provided to confirm the real-world applicability of the chosen methods. 

At this point we would also like to point out that the performance and reliability of the KIT19d were not only the product of the work conducted during the 2019 competition season. The development of our software stack was based on the software of the previous seasons. Instead of reinventing complete modules, improvements came often in small steps. This ensured a continuous performance improvement even with a small group of developers.

Last but not least we want to provide an outlook on topics we consider to be relevant for future improvements of the system. Extending the range of the perception system and lowering its inherent delays would allow for more accurate localization and control especially at high velocities. Regarding localization and mapping we consider SLAM methods, such as the EKF SLAM or GraphSLAM algorithms described in \cite{thrun2005robotics}, to be potential improvements upon our current implementation. Furthermore, recent developments from the field of model predictive control could allow for improvements of the motion planning and control modules. The two problems could be integrated in a single optimization problem, as presented in \cite{kabzan2019amz}. This would allow to simultaneously optimize the future trajectory and the lateral and longitudinal control inputs.


\section{Acknowledgements}
We would like to thank all present and past members of KA-RaceIng for working with us on this exceptional project. Successful participation in Formula Student Driverless requires much more than the software stack we outlined in this paper. Also, we want to thank our sponsors and all partners at the KIT for their continuous support without which we would not be able to turn our ideas into reality.